%% file: aaai24.tex
\documentclass[letterpaper]{article} 
\usepackage{aaai24}  
\usepackage{times}  
\usepackage{helvet}  
\usepackage{courier}  
\usepackage[hyphens]{url}  
\usepackage{graphicx} 
\usepackage{natbib}  
\usepackage{caption} 
\usepackage{algorithm}
\usepackage{algorithmic}

\usepackage{bbding}
\usepackage{bm,amsmath}
\usepackage[table,xcdraw]{xcolor}
\usepackage{multirow}
\usepackage{tabularx}
\usepackage{booktabs} 
\usepackage{amsthm}
\newcounter{myDefinition}
\newtheorem{definition}[myDefinition]{Definition}

\newcommand{\ie}{\emph{i.e., }}

\usepackage{newfloat}
\usepackage{listings}
\DeclareCaptionStyle{ruled}{labelfont=normalfont,labelsep=colon,strut=off} 
\floatstyle{ruled}
\newfloat{listing}{tb}{lst}{}
\floatname{listing}{Listing}
\nocopyright

\title{Text-to-Image Generation for Abstract Concepts}
\author {
    Jiayi Liao\textsuperscript{\rm 1}\equalcontrib \thanks{ This work is done during the internship in Microsoft.},
    Xu Chen\textsuperscript{\rm 2}\equalcontrib \thanks{Corresponding author.},
    Qiang Fu\textsuperscript{\rm 2},
    Lun Du\textsuperscript{\rm 2},\\
    Xiangnan He\textsuperscript{\rm 1},
    Xiang Wang\textsuperscript{\rm 1},
    Shi Han\textsuperscript{\rm 2},
    Dongmei Zhang\textsuperscript{\rm 2}
}
\affiliations {
    \textsuperscript{\rm 1} University of Science and Technology of China\\
    \textsuperscript{\rm 2} Microsoft\\
   ljy0ustc@mail.ustc.edu.cn, \{xu.chen, qifu, lun.du, shihan, dongmeiz\}@microsoft.com,\\ xiangnanhe@gmail.com, xiangwang@u.nus.edu
}

\begin{document}

\maketitle

\input{chapters/0_abstract}
\input{chapters/1_intro}
\input{chapters/2_related}
\input{chapters/3_method}
\input{chapters/4_exp}

\input{chapters/5_conclusion}

\input{aaai24.bbl}
\bibliographystyle{aaai24}
\bibliography{aaai24}
\clearpage

\end{document}

%% file: chapters/0_abstract.tex
\begin{abstract}
Recent years have witnessed the substantial progress of large-scale models across various domains, such as natural language processing and computer vision, facilitating the expression of concrete concepts. Unlike concrete concepts that are usually directly associated with physical objects, expressing abstract concepts through natural language requires considerable effort, which results from their intricate semantics and connotations. An alternative approach is to leverage images to convey rich visual information as a supplement. Nevertheless, existing Text-to-Image (T2I) models are primarily trained on concrete physical objects and tend to fail to visualize abstract concepts. Inspired by the three-layer artwork theory that identifies critical factors, \textbf{intent}, \textbf{object} and \textbf{form} during artistic creation, we propose a framework of \textbf{T}ext-to-\textbf{I}mage generation for \textbf{A}bstract \textbf{C}oncepts (\textbf{TIAC}). The abstract concept is clarified into a clear intent with a detailed definition to avoid ambiguity. LLMs then transform it into semantic-related physical objects, and the concept-dependent form is retrieved from an LLM-extracted form pattern set. Information from these three aspects will be integrated to generate prompts for T2I models via LLM. Evaluation results from human assessments and our newly designed metric \textbf{concept score} demonstrate the effectiveness of our framework in creating images that can sufficiently express abstract concepts.
\end{abstract}

%% file: chapters/1_intro.tex
\section{Introduction}
Concepts are cognitive representations that encapsulate ideas. The expression of concepts plays a pivotal role in communication, especially within the context of profound intellectual discourse. Recent advancements in large-scale models in the field of natural language processing and computer vision \cite{AIGC_survey1, AIGC_survey2, AIGC_survey3} have demonstrated remarkable performance in conveying concrete concepts. These concrete concepts are perceptible entities or occurrences and are often associated with physical objects, such as animals and planets. On the contrary, how to express abstract concepts that encompass rich and intricate connotations is less explored \cite{semantic_richness_of_abstract_concepts}. Abstract concepts serve a crucial role in accurately conveying philosophical thoughts, moral perspectives, and emotional states in everyday life. Moreover, they can present themes in a deeper and multi-dimensional manner in domains like art, literature, and music, rich with creativity and aesthetic value. On the other hand, abstract concepts are mentally constructed ideas that usually lack physical forms, and their definitions are often quite abstract as well. Therefore, when expressing abstract concepts through natural language, speakers encounter challenges in providing clear explanations, while recipients also face difficulties in understanding \cite{AbstractConcept}, leaving it a great barrier between human communications.

\begin{figure}[t]
    \centering
    \includegraphics[width=0.45\textwidth]{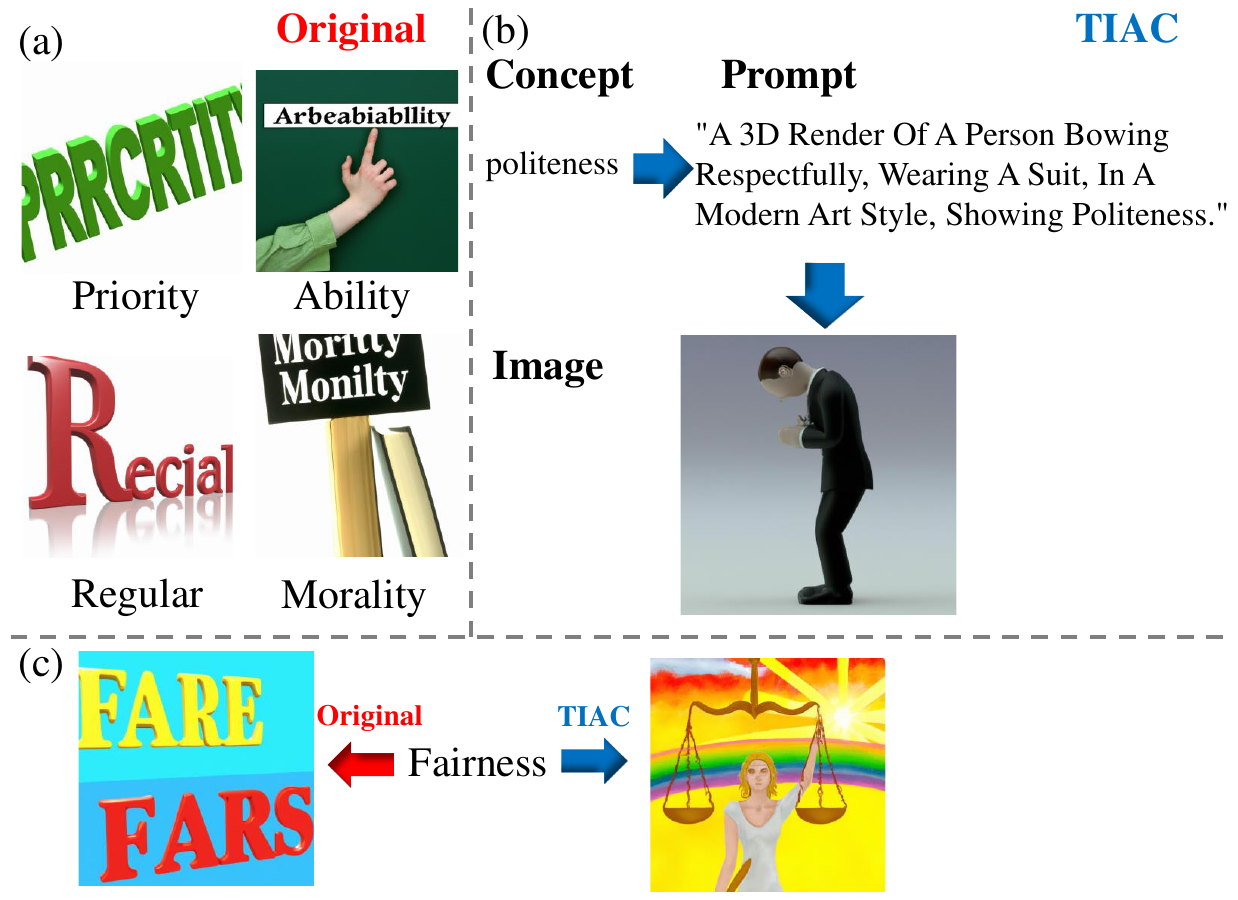}
    \vspace{-10pt}
    \caption{(a) Poor cases generated by DALL\textbullet E 2 for abstract concepts: ``priority'', ``ability'', ``regular'' and ``morality'', with concept names as inputs. (b) An illustration of ``politeness'' when introducing cooperation skills. The prompt describing how to draw an image about the concept of politeness is generated by TIAC. The image is generated with our prompt by DALL\textbullet E 2. (c) The left image is generated by directly taking the word ``fairness'' as input, and the right one is produced with the prompt from TIAC (LLM+PE). The right image consists of physical objects like a scale to express the concept of fairness.}
    \label{fig:comb_pics}
    \vspace{-15pt}
\end{figure}

\begin{figure*}[t]
    \centering
    \includegraphics[width=0.8\textwidth]{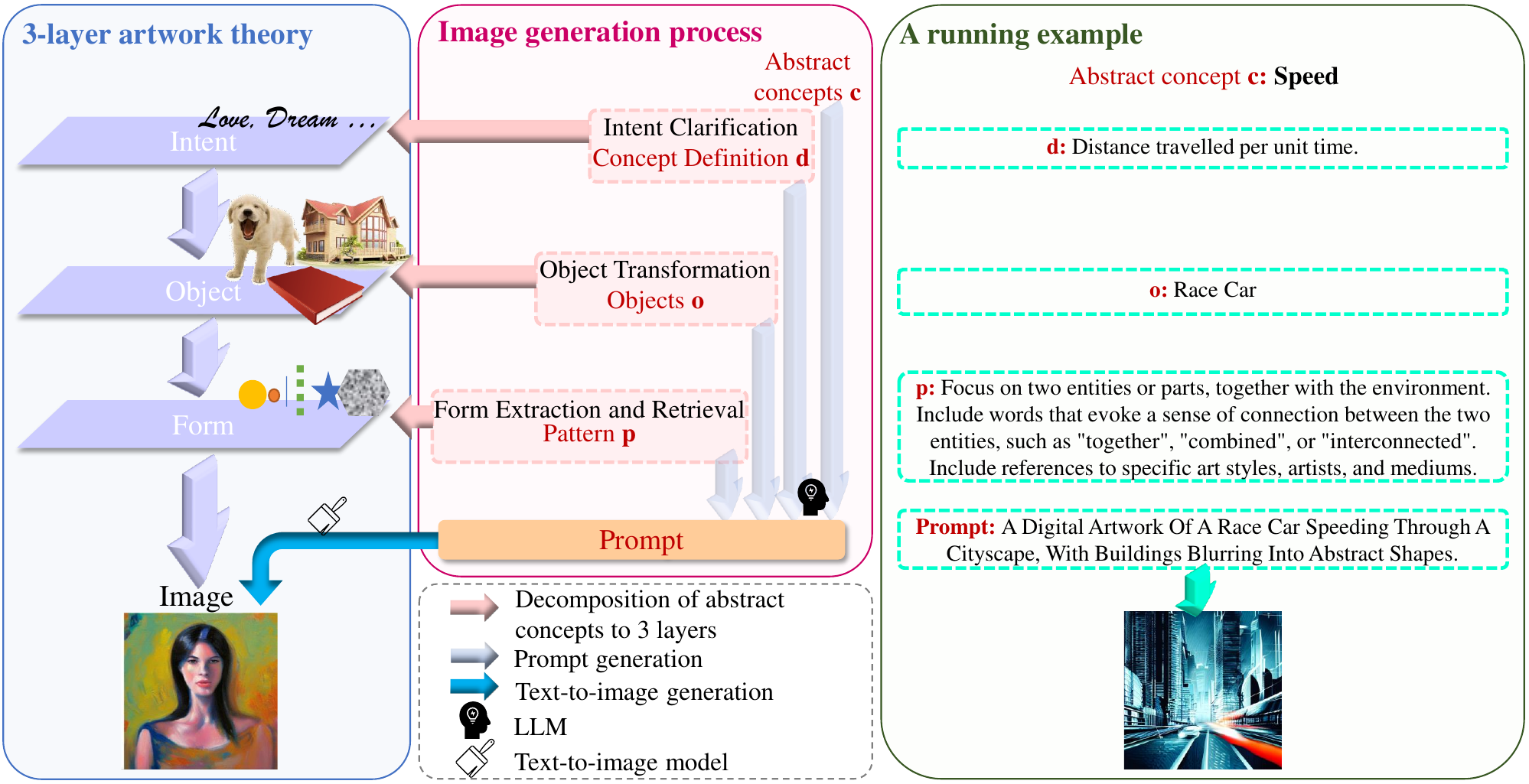}
    \vspace{-5pt}
    \caption{Framework of Image Generation for Abstract Concepts. The left section demonstrates the three layers considered when creating artwork: the intent layer, the object layer and the form layer. The middle part denotes that abstract concepts should be decomposed to 3 layers by Intent Clarification, Object Transformation and Form Extraction and Retrieval. Then we generate refined prompts with LLMs according to 3-layer information, and synthesize images with refined prompts using T2I models. The right section is a running example showing how TIAC depicts the concept ``speed''.}
    \label{fig:framework}
    \vspace{-15pt}
\end{figure*}

As an important channel of communication, conveying abstract concepts through vision can effectively alleviate the aforementioned challenges, enhancing the intuitive and vivid nature of their expression. 
For example, when using slides to introduce ``Cooperation Skills'' that include bullet points like ``Politeness'' and ``Fairness', incorporating relevant images to represent these abstract concepts can greatly improve communication efficiency: an image with a person bowing respectfully for ``Politeness'' and a person holding a scale for ``Fairness'' as shown in Figure \ref{fig:comb_pics} (b-c).

While it is of great potential to express abstract concepts through the visual channel, current text-to-image generation (T2I) models face obstacles in realizing this purpose. Existing T2I models such as Stable Diffusion \cite{StableDiffusion}, DALL\textbullet E \cite{Dalle}, Midjourney \cite{Midjourney}, NUWA \cite{NUWA} and Imagen \cite{Imagen} have achieved impressive improvements in generating realistic and eye-catching images with given input texts \cite{t2i_survey}. However, these models are primarily designed for concrete concepts, and high-quality datasets \cite{MSCOCO,Flickr,CUB,OxfordFlower} of text-image pairs used for their training only focus on physical objects, leading to the unsatisfactory generalization capabilities of T2I models for abstract concepts naturally \cite{Dalle,Imagen,GLIDE,Parti}.
When an abstract concept is directly inputted into a T2I model, it often generates images with distorted English letters resembling the input, as seen in the four images in Figure~\ref{fig:comb_pics} (a) and the left image of Figure~\ref{fig:comb_pics} (c). An analysis of prompts submitted by users on T2I servers \cite{PromptLogAnalysis} also reveals that images generated from vague and abstract prompts are often scored lower by users. 

To tackle the above challenges, we draw inspiration from a 3-layer artistic creation hierarchy \cite{ArtFundamentals, PromptLogAnalysis}, which is illustrated in Figure \ref{fig:framework}. The hierarchy is organized in a top-down manner, including: 
(1) \textbf{Intent} layer that reveals the high-level purpose the creator intends to express. (2) \textbf{Object} layer, which denotes physical objects and their spatial relationships. (3) \textbf{Form} layer that refers to the basic elements of artistic styles, such as line, color, shape, and texture, along with their arrangement. 
With the hierarchy in art, we can provide a new perspective to explain why T2I models perform worse for abstract concepts. When creating images for concrete concepts, information from the intent layer and the object layer are highly similar, i.e., what I think is what I will draw on the picture, and the corresponding form information is easy to obtain with plenty amount of images drawing physical objects. However, for abstract concepts, the connection between the concept and the information on the three layers is not obvious for current T2I models. Therefore, the key to effectively expressing abstract concepts lies in building connections to the three layers.


In this paper, we propose a framework of \textbf{T}ext-to-\textbf{I}mage Generation for \textbf{A}bstract \textbf{C}oncepts \textbf{TIAC} that aims to bridge the gap between human input abstract concepts and the generated images by leveraging the knowledge stored in LLMs and their comprehension ability. Specifically, TIAC links abstract concepts to nodes in WordNet so that the abstract concepts can be bonded to unambiguous definitions, which contributes to clarifying the user intents. Abstract concepts are then transformed by LLMs to related physical objects to represent their connotations. Additionally, the concept-dependent form patterns extracted from a prompt dataset are retrieved to enrich the information from the form layer. By integrating the above information from three layers, LLMs can generate prompts that tangibly describe abstract concepts, enabling T2I models to create satisfactory images.

Through conducting experiments on the abstract branch of WordNet, we compare different approaches and design a new metric called \textbf{concept score}. The results indicate that prompts generated using our framework facilitate effective visualization of abstract concepts. Furthermore, the concept score demonstrates better consistency with human preferences compared to existing metrics for assessing the alignment between abstract text inputs and generated images. Our framework, TIAC, optimizes prompts directly without necessitating model fine-tuning, making it adaptable to various T2I models.

The main contributions of our work are as follows:
\begin{itemize}
 \item We introduce a novel task of text-to-image generation for abstract concepts, aiming to fill the gap in abstract concept expression in the area of image generation.
 \item We design a framework TIAC to leverage LLMs to integrate the enriched information of abstract concepts in three layers. 
 \item We propose concept score, a new metric that is more aligned with human cognition for evaluating images generated for abstract concepts. Experimental results demonstrate the effectiveness of our TIAC in this novel task. 
\end{itemize}
It is worth noting that the image of ``politeness'' and ``fairness'' in Figure~\ref{fig:comb_pics} are both generated by our TIAC.

%% file: chapters/2_related.tex
\section{Related Work}
\subsubsection{Concept Expression in Image Generation.} 
In the exploration of image generation, there has been a recent upsurge in interest regarding concept expression.
For example, concept customization \cite{TextualInversion, DreamBooth, CustomDiffusion, ELITE} aims to integrate existing T2I models with new concepts.
However, these concepts are either specific objects like newly created objects "moongates" or customized objects in our daily lives like someone's pet dog.
Furthermore, concept disambiguation \cite{Ambiguity} also focuses on the syntactic equivocation inherent in human input, which leads to ambiguity concerning the referential relationships of physical objects, rather than delving into the subtle distinctions within abstract concepts. 
In general, current research in the field of image generation predominantly emphasizes the depiction of physical concepts rather than abstract ones.
Consequently, we aim to bridge the gap in the study of abstract concepts within this domain.

\subsubsection{Abstract Concepts in Computer Vision.}
One exemplary application of abstract concepts in computer vision is ad images.
Ad images are creative artworks that convey abstract concepts, incorporating a wealth of knowledge such as common-sense reasoning, cultural context, and symbolism.
Hence, tasks associated with ad images pose significant challenges for machines.
Several datasets \cite{ad-understanding-dataset,MetaCLUE,METMeme} and works of ads understanding \cite{ADVISE} and generation \cite{VisiBlends} have been proposed for visual ads.
The underlying idea of the paper mentioned before is combining objects in a logical manner to convey messages. 
This suggests that they also share the belief that the essence of abstract concept research lies in selecting objects and forms that can simultaneously express information.
Visual ads are a subset of our research, highlighting the potential application of abstract concept studies in the realm of artistic creation.
In addition, our applications also encompass illustrations of slides, decorative paintings, and more.

\subsubsection{Prompt Optimization for T2I Models.} 
While the performance of large-scale models driven by textual inputs is progressively advancing, the resource-intensive nature of training and fine-tuning these models has posed challenges for researchers to afford.
Consequently, improving the model input directly, known as prompt optimization, emerges as a commendable choice.
It can enhance image quality without altering the model structure or requiring extensive training and fine-tuning of T2I models.
This field is relatively new, resulting in a lack of comprehensive research.
 Experiences and insights of writing good T2I prompts manually are shared through blog posts and user guidebooks\cite{PromptTaxonomy,PromptTraveler, BestPrompt}.
Basic elements for a good T2I prompt and prompt terms (\ie modifiers) describing various perspectives of image style are summarized by a taxonomy survey \cite{PromptTaxonomy} and DALL\textbullet E 2 prompt book \cite{Dalle2PromptBook}.
As prompt optimization can be conducted in either text space or embedding space (i.e., soft tuning \cite{SoftPromptTuning}), some studies also train a prompt optimization model for soft tuning, which results in a high degree of coupling with the T2I model \cite{PromptOptimization}.
However, current T2I prompt optimization \cite{XPrompt} primarily aims to improve style and aesthetics of synthesized images.
In contrast, our research will prioritize the visual comprehension of concepts.

%% file: chapters/3_method.tex

\section{Preliminary}
\begin{definition}[Concept, concrete concept and abstract concept]
    Concepts are mental representations of coherent classes of entities \cite{AbstractConcept}; they can be divided into concrete concepts and abstract concepts. Concrete concepts are perceivable objects or occurrences, whereas abstract concepts are those that cannot be directly perceived through senses \cite{communicating_abstract_meaning,content_differences_abstract_concrete_concept}. 
\end{definition}
As intuitive examples, concrete concepts can be a tiger, a keyboard or a T-shirt, while abstract concepts can be dream, happiness or love.  Based on the above definition, the new task is further defined as: 
\begin{definition}[Text-to-image generation for abstract concept]
   Given a human text input that intends to express an abstract concept $\bm{c}$, the task of text-to-image generation for abstract concepts requires a mapping $\bm{f}$ to produce images that can reveal the meaning of $\bm{c}$.
    \begin{equation}
        \textbf{Image}=\bm{f}(\bm{c}).
        \label{eq:t2i-for-abstract-concepts}
\end{equation}
\end{definition}



\section{Method: TIAC}
To deal with this task, we propose TIAC which is inspired by the 3-layer artwork theory. The framework consists of 4 stages and is demonstrated in Figure~\ref{fig:framework}. (1) \textbf{Intent Clarification} stage aims to clarify the human intent. The input text will be linked to an existing entity in the WordNet knowledge base to retrieve its detailed definition as the definite intent. (2) \textbf{Object Transformation} stage is designed to decompose the abstract concept in the object layer. Here, concrete objects related to the intent will be obtained with the help of external knowledge and comprehension ability of LLMs. (3) \textbf{Form Extraction} stage will enrich the form information conditioned on the intent, which will be accessed through pre-extracted form patterns from a high-quality human-submitted prompt dataset. (4) \textbf{Prompt Generation and Image Generation} is the final stage and the above information will be integrated to generate prompts for T2I models so that more desirable images can be produced. Design in each stage will be elaborated on in this section.

\subsection{Intent Clarification}
\begin{figure*}[t]
    \centering
    \includegraphics[width=\textwidth]{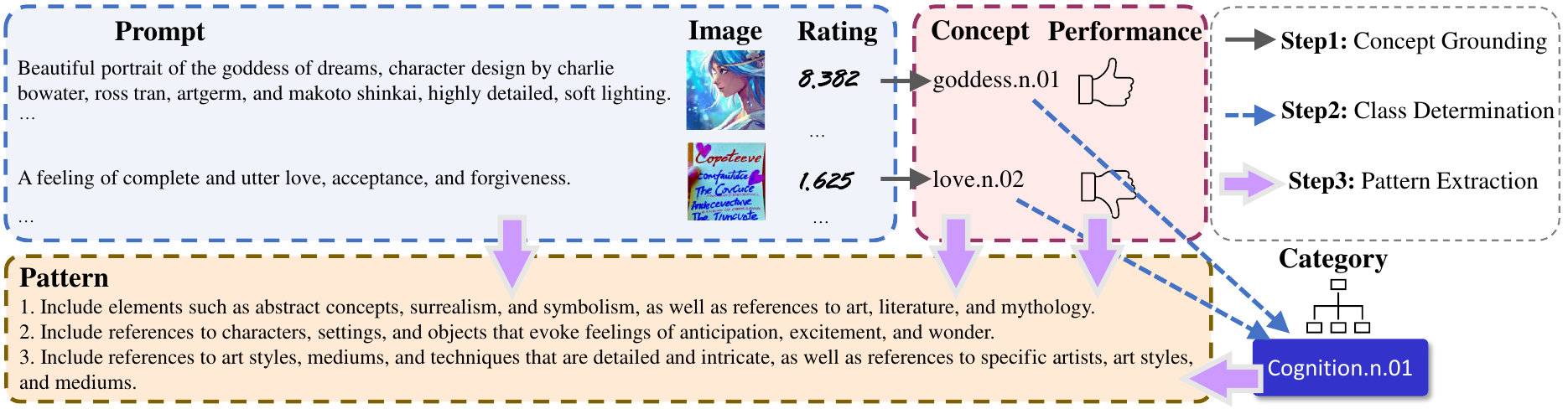}
    \vspace{-15pt}
    \caption{Details for 3 steps of Form Extraction in TIAC.}
    \label{fig:method}
    \vspace{-10pt}
\end{figure*}
When visualizing a human input abstract concept, its potential multiple connotations undoubtedly increase the difficulty of expression. For example, ``energy'' can mean (1) \textit{enterprising or ambitious drive} or (2) \textit{a thermodynamic quantity equivalent to the capacity of a physical system to do work} and so on.
Hence, it is imperative to clarify the precise semantics of the input abstract concept so that there is an exact drawing intent in the intent layer. To achieve this goal, an abstract concept is linked to a synset in WordNet \cite{wordnet} with a definite meaning. WordNet is a lexical database where semantically similar words are grouped into a set of cognitive synonyms called a synset; thus each synset represents a unique concept with corresponding definition in the database. Here, we focus on nouns in WordNet that are organized into a hierarchical tree with the root node of ``entity.n.01''\footnote{``entity.n.01'' is the notation of a synset in WordNet where ``entity'' denotes the synset name and ``n'' means it is a noun. Several synsets with the same name can be distinguished by the number like ``01''.}. There are two main branches under it: one is a subtree rooted at ``abstraction.n.06'' and the other rooted at ``physical\_entity.n.01.''. The former is the main focus of this paper as it represents abstract concepts.


Based on the subtree of abstract concepts $\bm{T}$, the intent clarification stage can be mathematically expressed as
\begin{equation}    \bm{d}=\bm{f}_{IC}(\bm{c}; \bm{T}).
    \label{eq:concept-linking}
\end{equation}
For each human input abstract concept $\bm{c}$, we link it to a specific node under the abstraction subtree $\bm{T}$ and retrieve the definition $\bm{d}$ of this node in WordNet. The intent clarification mapping $\bm{f}_{IC}$ builds a bridge between input abstract concepts and detailed drawing intent to mitigate potential ambiguity. 
Note that we assume the human input can be precisely mapped to a node in WordNet, whereas in the real scenario, it requires more efforts to determine the corresponding node in WordNet for input abstract concepts without other contexts. But the key idea is to identify the drawing intent of users, and we can achieve this by simply retrieving the definitions of all relevant WordNet nodes and asking users to decide from this candidate intent set. Overall, the definition $\bm{d}$ is now regarded as the exact intent in this task.

\subsection{Object Transformation}
The WordNet-based intent clarification alleviates the burden of ambiguous intent, but the intricate and abstract definition of abstract concepts can still be a barrier for this task. As seen in the case study, directly using the abstract concept or its definition as the input of T2I model both fail to yield satisfactory results, and the corresponding images are usually characters of input words with random noise instead of meaningful objects. On the other hand, there is a strong correlation between the intent layer and the object layer for concrete concepts; thus, they can be easily transformed into concrete objects and then illustrated by current T2I models. Hence, the critical step is to transform the abstract concept from the intent layer to the object layer. 

Actually, abstract concepts and concrete objects are not entirely irrelevant; in WordNet, the entity ``abstraction'' is defined as \textit{extracting common features from specific examples}, indicating that abstract concepts can be the summarization of properties or states of physical objects, or interactions between them. Therefore, concrete objects and their interactions can serve as instantiations of abstract concepts in reverse. 

To fulfill the transformation from the abstract concept to relevant physical objects or their interactions, we utilize LLMs for their knowledge and understanding ability. More specifically, LLMs have been trained with numerous corpus and should establish associations between concepts and relevant objects. With appropriate instructions, it can assist in the object transformation for given abstract concepts. This process can be formulated as: 
\begin{equation}
    \bm{o} = \bm{f}_{OT}(\bm{c}, \bm{d}; \bm{i}).
    \label{eq:object_transform}
\end{equation}
Here given an abstract concept $\bm{c}$ and its definition $\bm{d}$ as well as the object transformation instruction $\bm{i}$, the function $\bm{f}_{OT}$ (i.e., LLMs) is able to ground the human input to concrete objects $\bm{o}$.
As an example, the abstract concept ``shrinkage'' can be transformed to objects such as ``deflating balloon'', with $\bm{i}$ that employ words prompting LLMs to contain concrete objects which exemplify the meaning of concepts.

\subsection{Form Extraction and Retrieval}
Besides information from the object layer, incorporating information from the form layer also aids in expressing abstract concepts. The form layer depends on the intent layer, while LLMs may have difficulty generating form information for abstract concepts directly due to limited training data available about it. Hence, we introduce a dataset Simulacra Aesthetic Captions (SAC) \cite{SAC} to enhance LLMs in building the connection between intent and form. Moreover, abstract concepts in the same class are supposed to share common form information. Therefore, our approach involves initially extracting patterns pertaining to the form of a concept class. Subsequently, when a human input concept is provided, the corresponding form pattern is retrieved for the following usage. Details will be illustrated as follows.


\subsubsection{Form Extraction}
In form extraction, form patterns that describe the artistic properties of objects, how they are organized and the style of the picture are extracted with in-context learning from the SAC dataset. 
It consists of over 33k user-submitted T2I prompts and over 238k images generated from these prompts along with user ratings (1$\sim$10). In preprocessing, we assign the average rating as the score for prompts and remove prompts with fewer than 3 user ratings. The performance of prompts with scores below 3 is viewed as ``bad'' and above 8 is ``good''. 
Form extraction is conducted through three steps: concept grounding, class determination and pattern extraction, as shown in Figure \ref{fig:method}.


\noindent\underline{\textit{Concept Grounding.}} Human-submitted prompts contain other information besides the desired abstract concepts; thus, we ground the main concept in the prompt data $\bm{P}$ to a WordNet node $\bm{c}'$ by:
\begin{equation}    
\bm{c}'=\bm{f}_\text{ground}(\bm{P}; \bm{T}).
    \label{eq:concept-grounding-in-prompts}
\end{equation}
Here $\bm{f}_\text{ground}$ is an LLM that completes the tasks of (1) extracting the main concept of a prompt and ignoring artwork-related descriptions like pictures, images and so on; (2) grounding the concept to the most relevant synset in the abstract subtree $\bm{T}$ to obtain the intended abstract concept $\bm{c}'$.


\noindent\underline{\textit{Class Determination.}} Given the inadequate amount of prompt data for any single concepts while abstract concepts under the same ancestor share a common form pattern, we group the mapped nodes into different concept classes.
For an abstract concept $\bm{c}'$, its class can be obtained by
\begin{equation}
    \bm{C} = \bm{f}_\text{classify}(\bm{c}').
    \label{eq:concept-class-determination}
\end{equation}
$\bm{f}_\text{classify}$ determines the class by finding a common ancestor as the class label for a group of nodes and ensuring each class has enough nodes while maintaining distinctiveness. 

\noindent\underline{\textit{Pattern Extraction.}} Based on our pre-defined classes, the form pattern can be extracted with an LLM summarizing over a batch of prompts belonging to the same class in a contrastive manner as
\begin{equation}
    \bm{p} = \bm{f}_\text{extract}(\bm{C},\bm{D},\bm{P}_{good}, \bm{P}_{bad}).
\end{equation}
For a class $\bm{C}$, both good and bad prompts whose main concept is under $\bm{C}$ are put together. We instruct LLM $\bm{f}_\text{extract}$ to extract helpful form patterns from the good prompts $\bm{P}_{good}$ while avoiding harmful factors from bad ones $\bm{P}_{bad}$ based on the class and corresponding definition $\bm{D}$.
\subsubsection{Form Retrieve} 
With the extracted form pattern, we can acquire the form information for an intended abstract concept. Given that it has been mapped to $\bm{c}$ through intent clarification, the corresponding class $\bm{C}$ can be known following Eq.~\ref{eq:concept-class-determination} even if it does not appear in the SAC. Then the shared form pattern $\bm{p}$ can be retrieved from class $\bm{C}$, which has been extracted and stored during the form pattern extraction.

\subsection{Prompt Generation and Image Generation}
In cognitive psychology \cite{CognitivePsychology}, the cognitive processes of humans are defined as ``all mental processes by which the sensory input is transformed, reduced, elaborated, stored, recovered, and used."
In our framework, the process of intent clarification, object transformation as well as form extraction and retrieval are similar to the transformation, reduction, elaboration and storage of abstract concepts. Accordingly, prompt generation and image generation are the final steps, recovery and utilization.

From the above stages, We have obtained $\bm{d}$ from the intent layer, $\bm{o}$ from the object layer and $\bm{p}$ from the form layer. Furthermore, considering the benefits of in-context learning for LLMs, we select eight good prompts from the guidebook of DALL\textbullet E 2 \cite{Dalle2PromptBook} as few-shot examples $\bm{e}$. Also, the task description and the consideration of the token limit of downstream models are both incorporated. Hence, by utilizing the abilities of LLMs in information integration and content generation, the prompt for T2I generation model is obtained by
\begin{equation}  
    \textbf{Prompt}=\text{LLM}(\bm{c},\bm{d},\bm{o},\bm{p},\bm{e}).
    \label{eq:prompt}
\end{equation}
With T2I prompts from our designed framework, downstream T2I models are capable of generating images that can better express the intended abstract concepts:
\begin{equation}
    \textbf{Image}=\text{T2I-Model}(\textbf{Prompt}),
    \label{eq:image-generation}
\end{equation}
where T2I-Model can be any text-to-image generation model like Stable Diffusion 2 and DALL\textbullet E 2.

%% file: chapters/4_exp.tex
\section{Experiments}
\subsection{Experiment Settings}
\noindent\textbf{Dataset.}
We construct two datasets based on abstract concepts in WordNet with different scales. The small-scale one contains 57 abstract concepts and the large-scale one contains 3,400 abstract concepts.
(1) For each subclass with more than 100 nodes and under the seven classes on WordNet, we sample 100 abstract concepts to constitute the large-scale dataset.
(2) Based on the large-scale dataset, we further select three abstract concepts from each subclass whose number of prompts in SAC is over 10 to construct the small-scale dataset for human evaluation.

\noindent\textbf{Implementation details.}
T2I prompts generated from five different approaches are compared to verify the effectiveness: 
(1) \textbf{W} denotes taking \textbf{W}ords of the abstract concept as the prompt.
(2) \textbf{W+D} further concatenates the abstract concept name and its \textbf{D}efinition as the prompt.
(3) \textbf{LLM} means using the concept name and its definition as the input of \textbf{LLM} for prompt generation.
(4) \textbf{LLM+P} introduces information of the transformed objects and extracted form \textbf{P}atterns compared with \textbf{LLM}. 
(5) \textbf{LLM+PE} represents utilizing Eq.~\ref{eq:prompt} to obtain the prompts. Compared with LLM+P, few-shot \textbf{E}xamples $\bm{e}$ are added in LLM+PE.

We use GPT-3.5 \cite{GPT3} (text-davanci-003) as the LLM in our framework, and Stable Diffusion v2 \cite{StableDiffusion} (v2-inference and checkpoint of 512-base-ema) as the T2I model. Peformance of explicitly imposing object transformation is slightly off, so we use a more intuitive instruction in the last two baselines.

\begin{table}[t]
\centering
\begin{small}
\begin{tabular}{l |c|c|c|c}
\toprule
& HE & CS & IS & VS \\ 
\midrule
W & 1.86\tiny{$\pm$1.23}  & 0.91\tiny{$\pm$0.02}   & 2.58\tiny{$\pm$0.09} & 0.24\tiny{$\pm$0.00} \\ 
W+D & 2.60\tiny{$\pm$1.56}  & 1.05\tiny{$\pm$0.03}   & 2.66\tiny{$\pm$0.12}     & \textbf{0.29\tiny{$\pm$0.00}}                  \\ 
LLM & 3.34\tiny{$\pm$1.54}  & 1.14\tiny{$\pm$0.01} & \underline{2.75\tiny{$\pm$0.12}} & \underline{0.27\tiny{$\pm$0.00}} \\ 
LLM+P & \underline{3.52\tiny{$\pm$1.40}} & \underline{1.32\tiny{$\pm$0.01}} & 2.62\tiny{$\pm$0.11} & 0.25\tiny{$\pm$0.00} \\ 
LLM+PE & \textbf{3.80\tiny{$\pm$1.30}}                   & \textbf{1.50\tiny{$\pm$0.01}} & \textbf{2.76\tiny{$\pm$0.07}} & 0.25\tiny{$\pm$0.00} \\ 
\bottomrule
\end{tabular}
\vspace{-5pt}
\caption{
Evaluation Metrics on the Small-Scale Dataset. 
\textbf{Bold} and \underline{underline} indicate the best and the second best performance, respectively. HE: human evaluation, CS: concept score, IS: inception score, VS: visual-semantic similarity.
}
\label{tab:res_computed_small}
\end{small}
\vspace{-15pt}
\end{table}

\noindent\textbf{Evaluation.}
We conduct two types of evaluations: human evaluation and image-generation metrics evaluation.
(1) \textbf{\textit{Human Evaluation.}} We make a survey using images generated from small-scale dataset.
Given an abstract concept in the dataset, for each type of prompt, we generate three images for each prompt type and arrange them in a row.
The five types of images are randomly shuffled, with both the concept name and definition provided.
Respondents are asked to rate each row on a scale of 1 to 5 (1 is the worst and 5 is the best), indicating the perceived relevance between the images and the concept.
We collect feedback from three participants.
(2) \textbf{\textit{Image-generation Metrics Evaluation.}}
In the field of image generation, Inception Score \cite{IS} (IS) / Fréchet Inception Distance \cite{FID} (FID) are used for evaluating image quality and fidelity, and R-precision \cite{RPrecision} / Visual-Semantic similarity \cite{CLIP} (VS) are employed to measure the relevance between the image and the text.
We adopt IS and VS in our experiments for evaluation.

\subsection{Results on the small-scale dataset}
Experimental results of both human evaluation and image-generation metrics on the small-scale dataset are reported in Table \ref{tab:res_computed_small}.
Table \ref{tab:res_computed_small} demonstrates that LLM+P and LLM+PE achieve the top-2 results on human evaluation, showcasing significant improvements compared to taking raw human input or the concept definition as the prompts.
The detailed human assessment results are illustrated in Appendix.

\noindent\textbf{Discussion on Image-generation Metrics.} 
\begin{figure}[t]
    \centering
    \includegraphics[width=0.4\textwidth]{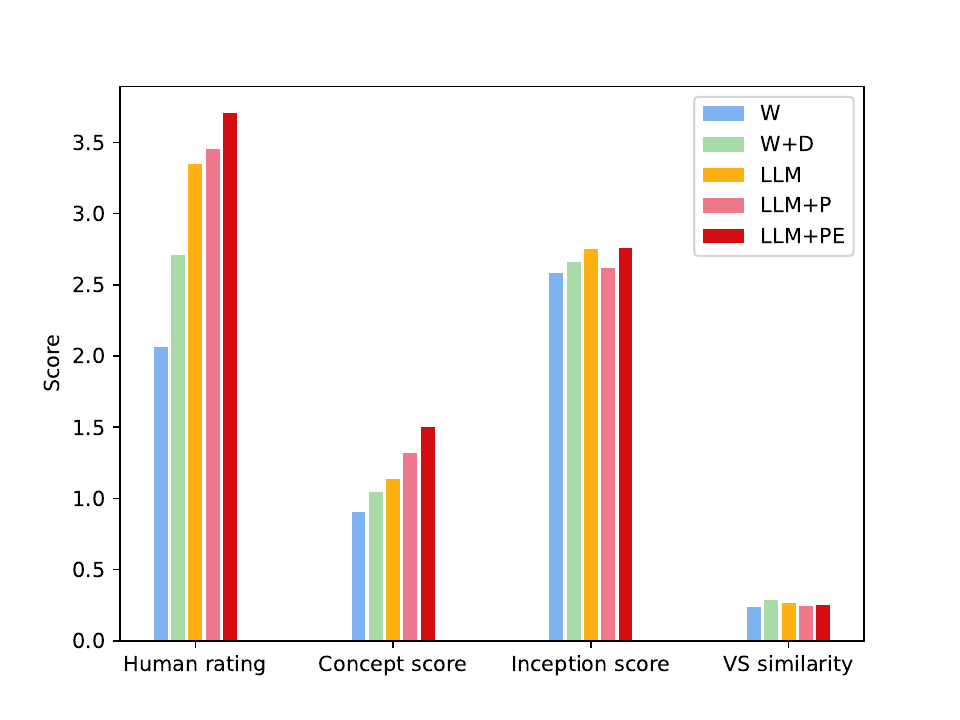}   \vspace{-10pt}\caption{Evaluation Metrics on small-scale dataset. Concept scores are more consistent with human ratings.}
    \label{fig:small_eval}
    \vspace{-15pt}
\end{figure}
Although our framework obtains the best performance in human assessment, Table~\ref{tab:res_computed_small} and Figure~\ref{fig:small_eval} also seem to demonstrate that LLM+PE and LLM+P are not so powerful on the image-generation metrics like IS and VS.
In other words, the machines seem to be against the conclusion that our design is remarkably better than simply inputting the desired abstract concept. What is the reason for this gap?
As a matter of fact, the task of image generation for abstract concepts is different from traditional image generation task that focuses on concrete concepts. For concrete 
concepts, there is a large number of training data and it is much easier to identify whether a concrete concept is correctly drawn on the image. 

\begin{table*}[t]
\centering
\begin{tabular}{c|c|c|c|c|c}
\toprule
\textbf{Category}           & \textbf{W} & \textbf{W+D} & \textbf{LLM} & \textbf{LLM+P}  & \textbf{LLM+PE}    \\ \midrule
\textbf{Attribute.n.02}     & 0.78±0.00     & 0.81±0.01    & 1.06±0.01    & \textbf{1.42±0.00} & 1.40±0.00              \\ 
\textbf{Cognition.n.01}     & 0.83±0.01  & 0.92±0.01    & 1.07±0.00       & 1.27±0.01       & \textbf{1.52±0.01} \\ 
\textbf{Communication.n.02} & 0.87±0.00     & 0.90±0.01     & 1.01±0.01    & 1.29±0.01       & \textbf{1.38±0.00}    \\ 
\textbf{Event.n.01}         & 0.92±0.00     & 1.01±0.01    & 1.17±0.01    & 1.38±0.00          & \textbf{1.50±0.00}     \\ 
\textbf{Group.n.01}         & 0.99±0.01  & 1.17±0.01    & 1.29±0.00       & 1.44±0.01       & \textbf{1.49±0.01} \\ 
\textbf{Measure.n.02}       & 0.98±0.01  & 1.12±0.01    & 1.21±0.00       & 1.26±0.01       & \textbf{1.31±0.00}    \\ 
\textbf{Relation.n.01}      & 0.85±0.01  & 0.93±0.01    & 1.13±0.01    & 1.13±0.02       & \textbf{1.34±0.01} \\ \midrule
\textbf{Average}      & 0.89±0.00  & 0.98±0.00    & 1.14±0.00    & 1.31±0.01       & \textbf{1.42±0.00} \\ \bottomrule
\end{tabular}
\caption{The Concept Score on the Large-Scale Dataset. \textbf{Bold} indicates the best performance. 0.00 means a very small deviation.}
\label{tab:simple_res_large}
\end{table*}


\begin{figure*}[t]
    \centering
    \includegraphics[width=0.7\textwidth]{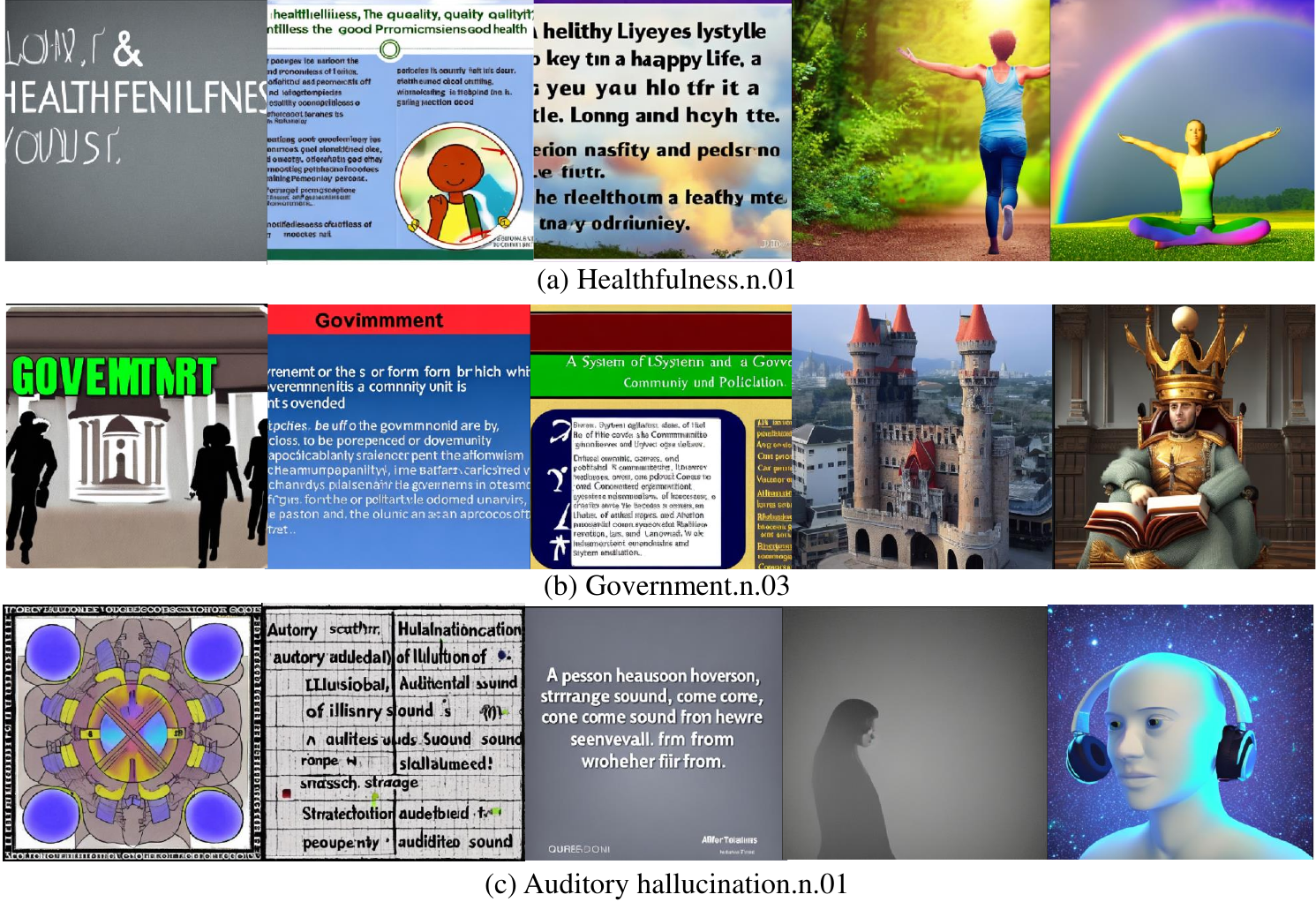}
    \vspace{-5pt}
    \caption{Case study. In each row, images from left to right are generated by five approaches: W, W+D, LLM, LLM+P, and LLM+PE. Healthfulness.n.01 means \textit{the quality of promoting good health}. Government.n.03 means \textit{the system or form by which a community or other political unit is governed}. Auditory hallucination.n.01 means \textit{illusory auditory perception of strange nonverbal sounds}. }
    \label{fig:image_cases}
     \vspace{-10pt}
\end{figure*}
However, to understand the underlying abstract concepts, it requires deeper processing progress on the combinations of objects and forms (recalling viewing pieces of art in the exhibition), which is beyond the existing image-generation metrics. Actually, the calculation of the IS utilizes Inception Network to classify the images into the classes in ImageNet 1k, and the computation of VS encodes images and input text into a shared latent space using a pre-trained encoder (here we use CLIP for both image encoding and text encoding), and then calculates the distance between two vectors to measure the alignment between image and text. They are both designed for concrete objects, leaving the emergency of designing an evaluation metric for abstract objects.

To this end, we introduce the Aesthetic score \cite{LaionAesthetics} that is proposed to evaluate a generated image from the perspective of art considering abstraction to some extent. Hence, we consider both factors and take the aesthetic score as a coefficient for VS similarity as the measure of alignment between texts and images with abstraction:
\begin{equation}
    \textbf{Concept score}=\textbf{VS similarity} \times \textbf{Aesthetic score}
\end{equation}

By revisiting Figure ~\ref{fig:small_eval}, we can find that the concept score is more consistent with human preference. Hence, it enables us to evaluate the five approaches on the large-scale dataset.

\subsection{Results on the large-scale dataset}
We further validate the effectiveness of different prompts on large-scale dataset with concept score. The calculation is conducted over 3400 abstract concepts from 34 subclasses under the seven classes. The statistical results on the seven classes are organized in Table~\ref{tab:simple_res_large} and detailed information is provided in Appendix due to the page limit. Similarly, LLM+P and LLM+PE generally obtain the highest concept score, which illustrates that even being evaluated on a more diverse dataset with more abstract concepts, our design can still outperform other baselines and generate better images.

\subsection{Case Study}
We randomly select three abstract concepts and generate images with five different types of prompts to directly demonstrate the performance. As shown in Figure ~\ref{fig:image_cases}, LLM+P and LLM+PE generated images are much more meaningful and beautiful than the left ones. Concretely, LLM+P and LLM+PE convey the concept of ``healthfulness'' through running in the green forest or practicing yoga on the grass, and express ``government'' through a magnificent castle and a king sitting on the throne in the palace. For ``auditory hallucination'', LLM+P seems to produce a less relevant image, while LLM+PE generates an image where a person is wearing headphones with an ethereality background that is a little bit close to the concept.

%% file: chapters/5_conclusion.tex
\section{Conclusion}
In this paper, we study a new task text-to-image generation for abstract concepts.
The difficulty mainly arises from the intricate connotations associated with abstract concepts, making them challenging to explain and understand.
Hence, we propose a framework called TIAC, leveraging the comprehension and generation abilities of LLMs. 
TIAC first enriches information from three layers by clarifying the user intent based on WordNet, transforming abstract intents into objects, and retrieving form patterns extracted from SAC dataset.
By integrating information from three layers, LLMs can construct effective T2I prompts, resulting in the generation of better images.
We conduct both human assessment and metrics evaluation, and design a new metric, concept score.
The evaluation result and case study show our superiority in the task over baselines.


%% file: aaai24.bbl
\begin{thebibliography}{47}
\providecommand{\natexlab}[1]{#1}

\bibitem[{Akula et~al.(2022)Akula, Driscoll, Narayana, Changpinyo, Jia, Damle, Pruthi, Basu, Guibas, Freeman, Li, and Jampani}]{MetaCLUE}
Akula, A.~R.; Driscoll, B.; Narayana, P.; Changpinyo, S.; Jia, Z.; Damle, S.; Pruthi, G.; Basu, S.; Guibas, L.~J.; Freeman, W.~T.; Li, Y.; and Jampani, V. 2022.
\newblock MetaCLUE: Towards Comprehensive Visual Metaphors Research.
\newblock \emph{CoRR}, abs/2212.09898.

\bibitem[{Brown et~al.(2020)Brown, Mann, Ryder, Subbiah, Kaplan, Dhariwal, Neelakantan, Shyam, Sastry, Askell, Agarwal, Herbert{-}Voss, Krueger, Henighan, Child, Ramesh, Ziegler, Wu, Winter, Hesse, Chen, Sigler, Litwin, Gray, Chess, Clark, Berner, McCandlish, Radford, Sutskever, and Amodei}]{GPT3}
Brown, T.~B.; Mann, B.; Ryder, N.; Subbiah, M.; Kaplan, J.; Dhariwal, P.; Neelakantan, A.; Shyam, P.; Sastry, G.; Askell, A.; Agarwal, S.; Herbert{-}Voss, A.; Krueger, G.; Henighan, T.; Child, R.; Ramesh, A.; Ziegler, D.~M.; Wu, J.; Winter, C.; Hesse, C.; Chen, M.; Sigler, E.; Litwin, M.; Gray, S.; Chess, B.; Clark, J.; Berner, C.; McCandlish, S.; Radford, A.; Sutskever, I.; and Amodei, D. 2020.
\newblock Language Models are Few-Shot Learners.
\newblock In \emph{NeurIPS}.

\bibitem[{Cao et~al.(2023)Cao, Li, Liu, Yan, Dai, Yu, and Sun}]{AIGC_survey1}
Cao, Y.; Li, S.; Liu, Y.; Yan, Z.; Dai, Y.; Yu, P.~S.; and Sun, L. 2023.
\newblock A Comprehensive Survey of AI-Generated Content {(AIGC):} {A} History of Generative {AI} from {GAN} to ChatGPT.
\newblock \emph{CoRR}, abs/2303.04226.

\bibitem[{Chilton, Petridis, and Agrawala(2019)}]{VisiBlends}
Chilton, L.~B.; Petridis, S.; and Agrawala, M. 2019.
\newblock VisiBlends: {A} Flexible Workflow for Visual Blends.
\newblock In \emph{{CHI}}, 172. {ACM}.

\bibitem[{Gal et~al.(2022)Gal, Alaluf, Atzmon, Patashnik, Bermano, Chechik, and Cohen{-}Or}]{TextualInversion}
Gal, R.; Alaluf, Y.; Atzmon, Y.; Patashnik, O.; Bermano, A.~H.; Chechik, G.; and Cohen{-}Or, D. 2022.
\newblock An Image is Worth One Word: Personalizing Text-to-Image Generation using Textual Inversion.
\newblock \emph{CoRR}, abs/2208.01618.

\bibitem[{Ge et~al.(2022)Ge, Hu, Dong, Mao, Xia, Wang, Chen, and Wei}]{XPrompt}
Ge, T.; Hu, J.; Dong, L.; Mao, S.; Xia, Y.; Wang, X.; Chen, S.; and Wei, F. 2022.
\newblock Extensible Prompts for Language Models.
\newblock \emph{CoRR}, abs/2212.00616.

\bibitem[{Hao et~al.(2022)Hao, Chi, Dong, and Wei}]{PromptOptimization}
Hao, Y.; Chi, Z.; Dong, L.; and Wei, F. 2022.
\newblock Optimizing Prompts for Text-to-Image Generation.
\newblock \emph{CoRR}, abs/2212.09611.

\bibitem[{Heusel et~al.(2017)Heusel, Ramsauer, Unterthiner, Nessler, and Hochreiter}]{FID}
Heusel, M.; Ramsauer, H.; Unterthiner, T.; Nessler, B.; and Hochreiter, S. 2017.
\newblock GANs Trained by a Two Time-Scale Update Rule Converge to a Local Nash Equilibrium.
\newblock In \emph{{NIPS}}, 6626--6637.

\bibitem[{Hussain et~al.(2017)Hussain, Zhang, Zhang, Ye, Thomas, Agha, Ong, and Kovashka}]{ad-understanding-dataset}
Hussain, Z.; Zhang, M.; Zhang, X.; Ye, K.; Thomas, C.; Agha, Z.; Ong, N.; and Kovashka, A. 2017.
\newblock Automatic Understanding of Image and Video Advertisements.
\newblock In \emph{{CVPR}}, 1100--1110. {IEEE} Computer Society.

\bibitem[{Katja Wiemer-Hastings and Xu(2005)}]{content_differences_abstract_concrete_concept}
Katja Wiemer-Hastings, K.; and Xu, X. 2005.
\newblock Content differences for abstract and concrete concepts.
\newblock \emph{Cognitive science}, 29(5): 719--736.

\bibitem[{Kumari et~al.(2022)Kumari, Zhang, Zhang, Shechtman, and Zhu}]{CustomDiffusion}
Kumari, N.; Zhang, B.; Zhang, R.; Shechtman, E.; and Zhu, J. 2022.
\newblock Multi-Concept Customization of Text-to-Image Diffusion.
\newblock \emph{CoRR}, abs/2212.04488.

\bibitem[{Lester, Al{-}Rfou, and Constant(2021)}]{SoftPromptTuning}
Lester, B.; Al{-}Rfou, R.; and Constant, N. 2021.
\newblock The Power of Scale for Parameter-Efficient Prompt Tuning.
\newblock In \emph{{EMNLP} {(1)}}, 3045--3059. Association for Computational Linguistics.

\bibitem[{Lin et~al.(2014)Lin, Maire, Belongie, Hays, Perona, Ramanan, Doll{\'{a}}r, and Zitnick}]{MSCOCO}
Lin, T.; Maire, M.; Belongie, S.~J.; Hays, J.; Perona, P.; Ramanan, D.; Doll{\'{a}}r, P.; and Zitnick, C.~L. 2014.
\newblock Microsoft {COCO:} Common Objects in Context.
\newblock In \emph{{ECCV} {(5)}}, volume 8693 of \emph{Lecture Notes in Computer Science}, 740--755. Springer.

\bibitem[{Mehrabi et~al.(2022)Mehrabi, Goyal, Verma, Dhamala, Kumar, Hu, Chang, Zemel, Galstyan, and Gupta}]{Ambiguity}
Mehrabi, N.; Goyal, P.; Verma, A.; Dhamala, J.; Kumar, V.; Hu, Q.; Chang, K.; Zemel, R.~S.; Galstyan, A.; and Gupta, R. 2022.
\newblock Is the Elephant Flying? Resolving Ambiguities in Text-to-Image Generative Models.
\newblock \emph{CoRR}, abs/2211.12503.

\bibitem[{midjourney.com(Retrieved on 5/5/2023)}]{Midjourney}
midjourney.com. Retrieved on 5/5/2023.
\newblock Midjourney.

\bibitem[{Miller(1995)}]{wordnet}
Miller, G.~A. 1995.
\newblock WordNet: {A} Lexical Database for English.
\newblock \emph{Commun. {ACM}}, 38(11): 39--41.

\bibitem[{Neisser(2014)}]{CognitivePsychology}
Neisser, U. 2014.
\newblock \emph{Cognitive psychology: Classic edition}.
\newblock Psychology press.

\bibitem[{Nichol et~al.(2022)Nichol, Dhariwal, Ramesh, Shyam, Mishkin, McGrew, Sutskever, and Chen}]{GLIDE}
Nichol, A.~Q.; Dhariwal, P.; Ramesh, A.; Shyam, P.; Mishkin, P.; McGrew, B.; Sutskever, I.; and Chen, M. 2022.
\newblock {GLIDE:} Towards Photorealistic Image Generation and Editing with Text-Guided Diffusion Models.
\newblock In \emph{{ICML}}, volume 162 of \emph{Proceedings of Machine Learning Research}, 16784--16804. {PMLR}.

\bibitem[{Nilsback and Zisserman(2008)}]{OxfordFlower}
Nilsback, M.; and Zisserman, A. 2008.
\newblock Automated Flower Classification over a Large Number of Classes.
\newblock In \emph{{ICVGIP}}, 722--729. {IEEE} Computer Society.

\bibitem[{Ocvirk et~al.(1968)Ocvirk, Stinson, Wigg, Bone, and Cayton}]{ArtFundamentals}
Ocvirk, O.~G.; Stinson, R.~E.; Wigg, P.~R.; Bone, R.~O.; and Cayton, D.~L. 1968.
\newblock \emph{Art fundamentals: Theory and practice}.
\newblock WC Brown Company.

\bibitem[{Oppenlaender(2022)}]{PromptTaxonomy}
Oppenlaender, J. 2022.
\newblock A Taxonomy of Prompt Modifiers for Text-to-Image Generation.
\newblock \emph{arXiv preprint arXiv:2204.13988}.

\bibitem[{Parsons(2022)}]{Dalle2PromptBook}
Parsons, G. 2022.
\newblock The DALL·E 2 Prompt Book.

\bibitem[{Pavlichenko and Ustalov(2022)}]{BestPrompt}
Pavlichenko, N.; and Ustalov, D. 2022.
\newblock Best Prompts for Text-to-Image Models and How to Find Them.
\newblock \emph{CoRR}, abs/2209.11711.

\bibitem[{Pressman, Crowson, and Contributors(2022)}]{SAC}
Pressman, J.~D.; Crowson, K.; and Contributors, S.~C. 2022.
\newblock Simulacra Aesthetic Captions.
\newblock Technical Report Version 1.0, Stability AI.
\newblock \ url { https://github.com/JD-P/simulacra-aesthetic-captions }.

\bibitem[{Radford et~al.(2021)Radford, Kim, Hallacy, Ramesh, Goh, Agarwal, Sastry, Askell, Mishkin, Clark, Krueger, and Sutskever}]{CLIP}
Radford, A.; Kim, J.~W.; Hallacy, C.; Ramesh, A.; Goh, G.; Agarwal, S.; Sastry, G.; Askell, A.; Mishkin, P.; Clark, J.; Krueger, G.; and Sutskever, I. 2021.
\newblock Learning Transferable Visual Models From Natural Language Supervision.
\newblock In \emph{{ICML}}, volume 139 of \emph{Proceedings of Machine Learning Research}, 8748--8763. {PMLR}.

\bibitem[{Ramesh et~al.(2022)Ramesh, Dhariwal, Nichol, Chu, and Chen}]{Dalle}
Ramesh, A.; Dhariwal, P.; Nichol, A.; Chu, C.; and Chen, M. 2022.
\newblock Hierarchical Text-Conditional Image Generation with {CLIP} Latents.
\newblock \emph{CoRR}, abs/2204.06125.

\bibitem[{Recchia and Jones(2012)}]{semantic_richness_of_abstract_concepts}
Recchia, G.; and Jones, M.~N. 2012.
\newblock The semantic richness of abstract concepts.
\newblock \emph{Frontiers in human neuroscience}, 6: 315.

\bibitem[{Rombach et~al.(2022)Rombach, Blattmann, Lorenz, Esser, and Ommer}]{StableDiffusion}
Rombach, R.; Blattmann, A.; Lorenz, D.; Esser, P.; and Ommer, B. 2022.
\newblock High-Resolution Image Synthesis with Latent Diffusion Models.
\newblock In \emph{{CVPR}}, 10674--10685. {IEEE}.

\bibitem[{Ruiz et~al.(2022)Ruiz, Li, Jampani, Pritch, Rubinstein, and Aberman}]{DreamBooth}
Ruiz, N.; Li, Y.; Jampani, V.; Pritch, Y.; Rubinstein, M.; and Aberman, K. 2022.
\newblock DreamBooth: Fine Tuning Text-to-Image Diffusion Models for Subject-Driven Generation.
\newblock \emph{CoRR}, abs/2208.12242.

\bibitem[{Saharia et~al.(2022)Saharia, Chan, Saxena, Li, Whang, Denton, Ghasemipour, Ayan, Mahdavi, Lopes, Salimans, Ho, Fleet, and Norouzi}]{Imagen}
Saharia, C.; Chan, W.; Saxena, S.; Li, L.; Whang, J.; Denton, E.; Ghasemipour, S. K.~S.; Ayan, B.~K.; Mahdavi, S.~S.; Lopes, R.~G.; Salimans, T.; Ho, J.; Fleet, D.~J.; and Norouzi, M. 2022.
\newblock Photorealistic Text-to-Image Diffusion Models with Deep Language Understanding.
\newblock \emph{CoRR}, abs/2205.11487.

\bibitem[{Salimans et~al.(2016)Salimans, Goodfellow, Zaremba, Cheung, Radford, and Chen}]{IS}
Salimans, T.; Goodfellow, I.~J.; Zaremba, W.; Cheung, V.; Radford, A.; and Chen, X. 2016.
\newblock Improved Techniques for Training GANs.
\newblock In \emph{{NIPS}}, 2226--2234.

\bibitem[{Schuhmann(2022)}]{LaionAesthetics}
Schuhmann, C. 2022.
\newblock LAION-AESTHETICS.

\bibitem[{Schwanenflugel(2013)}]{AbstractConcept}
Schwanenflugel, P.~J. 2013.
\newblock \emph{The psychology of word meanings}.
\newblock Psychology Press.

\bibitem[{Smith(2022)}]{PromptTraveler}
Smith, E. 2022.
\newblock A Traveler’s Guide to the Latent Space.

\bibitem[{Wah et~al.(2011)Wah, Branson, Welinder, Perona, and Belongie}]{CUB}
Wah, C.; Branson, S.; Welinder, P.; Perona, P.; and Belongie, S. 2011.
\newblock The caltech-ucsd birds-200-2011 dataset.

\bibitem[{Wei et~al.(2023)Wei, Zhang, Ji, Bai, Zhang, and Zuo}]{ELITE}
Wei, Y.; Zhang, Y.; Ji, Z.; Bai, J.; Zhang, L.; and Zuo, W. 2023.
\newblock {ELITE:} Encoding Visual Concepts into Textual Embeddings for Customized Text-to-Image Generation.
\newblock \emph{CoRR}, abs/2302.13848.

\bibitem[{Wu et~al.(2022)Wu, Liang, Ji, Yang, Fang, Jiang, and Duan}]{NUWA}
Wu, C.; Liang, J.; Ji, L.; Yang, F.; Fang, Y.; Jiang, D.; and Duan, N. 2022.
\newblock N{\"{U}}WA: Visual Synthesis Pre-training for Neural visUal World creAtion.
\newblock In \emph{{ECCV} {(16)}}, volume 13676 of \emph{Lecture Notes in Computer Science}, 720--736. Springer.

\bibitem[{Wu et~al.(2023)Wu, Gan, Chen, Wan, and Lin}]{AIGC_survey3}
Wu, J.; Gan, W.; Chen, Z.; Wan, S.; and Lin, H. 2023.
\newblock AI-Generated Content {(AIGC):} {A} Survey.
\newblock \emph{CoRR}, abs/2304.06632.

\bibitem[{Xie et~al.(2023)Xie, Pan, Ma, Jie, and Mei}]{PromptLogAnalysis}
Xie, Y.; Pan, Z.; Ma, J.; Jie, L.; and Mei, Q. 2023.
\newblock A Prompt Log Analysis of Text-to-Image Generation Systems.
\newblock In \emph{{WWW}}, 3892--3902. {ACM}.

\bibitem[{Xu et~al.(2022)Xu, Li, Zheng, Naseriparsa, Zhao, Lin, and Xia}]{METMeme}
Xu, B.; Li, T.; Zheng, J.; Naseriparsa, M.; Zhao, Z.; Lin, H.; and Xia, F. 2022.
\newblock MET-Meme: {A} Multimodal Meme Dataset Rich in Metaphors.
\newblock In \emph{{SIGIR}}, 2887--2899. {ACM}.

\bibitem[{Xu et~al.(2018)Xu, Zhang, Huang, Zhang, Gan, Huang, and He}]{RPrecision}
Xu, T.; Zhang, P.; Huang, Q.; Zhang, H.; Gan, Z.; Huang, X.; and He, X. 2018.
\newblock AttnGAN: Fine-Grained Text to Image Generation With Attentional Generative Adversarial Networks.
\newblock In \emph{{CVPR}}, 1316--1324. Computer Vision Foundation / {IEEE} Computer Society.

\bibitem[{Ye and Kovashka(2018)}]{ADVISE}
Ye, K.; and Kovashka, A. 2018.
\newblock {ADVISE:} Symbolism and External Knowledge for Decoding Advertisements.
\newblock In \emph{{ECCV} {(15)}}, volume 11219 of \emph{Lecture Notes in Computer Science}, 868--886. Springer.

\bibitem[{Young et~al.(2014)Young, Lai, Hodosh, and Hockenmaier}]{Flickr}
Young, P.; Lai, A.; Hodosh, M.; and Hockenmaier, J. 2014.
\newblock From image descriptions to visual denotations: New similarity metrics for semantic inference over event descriptions.
\newblock \emph{Trans. Assoc. Comput. Linguistics}, 2: 67--78.

\bibitem[{Yu et~al.(2022)Yu, Xu, Koh, Luong, Baid, Wang, Vasudevan, Ku, Yang, Ayan, Hutchinson, Han, Parekh, Li, Zhang, Baldridge, and Wu}]{Parti}
Yu, J.; Xu, Y.; Koh, J.~Y.; Luong, T.; Baid, G.; Wang, Z.; Vasudevan, V.; Ku, A.; Yang, Y.; Ayan, B.~K.; Hutchinson, B.; Han, W.; Parekh, Z.; Li, X.; Zhang, H.; Baldridge, J.; and Wu, Y. 2022.
\newblock Scaling Autoregressive Models for Content-Rich Text-to-Image Generation.
\newblock \emph{CoRR}, abs/2206.10789.

\bibitem[{Zdrazilova, Sidhu, and Pexman(2018)}]{communicating_abstract_meaning}
Zdrazilova, L.; Sidhu, D.~M.; and Pexman, P.~M. 2018.
\newblock Communicating abstract meaning: concepts revealed in words and gestures.
\newblock \emph{Philosophical Transactions of the Royal Society B: Biological Sciences}, 373(1752): 20170138.

\bibitem[{Zhang et~al.(2023{\natexlab{a}})Zhang, Zhang, Zhang, and Kweon}]{t2i_survey}
Zhang, C.; Zhang, C.; Zhang, M.; and Kweon, I.~S. 2023{\natexlab{a}}.
\newblock Text-to-image Diffusion Models in Generative {AI:} {A} Survey.
\newblock \emph{CoRR}, abs/2303.07909.

\bibitem[{Zhang et~al.(2023{\natexlab{b}})Zhang, Zhang, Zheng, Qiao, Li, Zhang, Dam, Thwal, Tun, Huy, Kim, Bae, Lee, Yang, Shen, Kweon, and Hong}]{AIGC_survey2}
Zhang, C.; Zhang, C.; Zheng, S.; Qiao, Y.; Li, C.; Zhang, M.; Dam, S.~K.; Thwal, C.~M.; Tun, Y.~L.; Huy, L.~L.; Kim, D.~U.; Bae, S.; Lee, L.; Yang, Y.; Shen, H.~T.; Kweon, I.~S.; and Hong, C.~S. 2023{\natexlab{b}}.
\newblock A Complete Survey on Generative {AI} {(AIGC):} Is ChatGPT from {GPT-4} to {GPT-5} All You Need?
\newblock \emph{CoRR}, abs/2303.11717.

\end{thebibliography}
